\documentclass[11pt]{article}

\usepackage[preprint]{acl}

\usepackage{times}
\usepackage{latexsym}
\usepackage{amsmath, amssymb}

\usepackage[T1]{fontenc}
\usepackage[utf8]{inputenc}
\usepackage{breqn}

\usepackage{microtype}

\usepackage{inconsolata}

\usepackage{graphicx}

\usepackage{amsmath}    % 数学符号，如 \downarrow
\usepackage{booktabs}   % 三线表 \toprule \midrule \bottomrule
\usepackage{textcomp}   % 保证 \textsubscript 可用
\usepackage{graphicx}   % 插入图片
\usepackage{subcaption} % 子图环境

\title{TrimTokenator: Towards Adaptive Visual Token Pruning for Large Multimodal Models}

\author{\bf Hao Zhang$^{1}$\thanks{These authors contribute equally to this work.}, Mengsi Lyu$^{1}$\footnotemark[1], Chenrui He$^{1}$, Yulong Ao$^1$\thanks{Corresponding author.},  Yonghua Lin$^1$\footnotemark[2]\\
\normalsize{}$^1$Beijing Academy of Artificial Intelligence (BAAI)
}

\begin{document}
\maketitle
\begin{abstract}
Large Multimodal Models (LMMs) have achieved significant success across various tasks. These models usually encode visual inputs into dense token sequences, which are then concatenated with textual tokens and jointly processed by a language model. However, the increased token count substantially raises computational and memory costs during inference. Token pruning has emerged as a promising approach to address this issue. Existing token pruning methods often rely on costly calibration or suboptimal importance metrics, leading to redundant retained tokens. In this paper, we analyze the redundancy differences between visual and textual tokens and propose pruning exclusively on visual tokens. Based on this, we propose a visual token pruning strategy that explicitly preserves both cross-modal alignment and intra-modal informational diversity. We introduce a mutual information-based token pruning strategy that removes visual tokens semantically misaligned with textual tokens, effectively preserving the alignment between the visual and textual modalities. To further improve the representational quality of the retained tokens, we additionally prune redundant visual tokens by maximizing the expected pairwise distances in the embedding space, which is solved efficiently with a greedy algorithm. Extensive experiments demonstrate that our method maintains strong performance while reducing tokens by 88.9\% on models such as LLaVA-1.5-7B and LLaVA-NEXT-7B, resulting in a 56.7\% improvement in inference speed.

\end{abstract}

\section{Introduction}
Large Multimodal Models (LMMs) \cite{bai2025qwen25vl,team2025kimivl,zhu2025internvl3,liu2024llava15,lin2023videollava} have substantially enhanced the reasoning capabilities of Large Language Models (LLMs) \cite{brown2020language, touvron2023open, chiang2023vicuna,zhu2023a,li2023,zhang2023,huang2023,wang2023} by enabling joint processing of multimodal inputs such as images and texts. Typically, visual inputs are encoded into dense token sequences via a vision encoder and concatenated with textual tokens for unified processing by the language model. However, the resulting token sequences often reach thousands in length, leading to significant computational and memory overhead due to the quadratic complexity of self-attention with respect to sequence length \cite{vaswani2017attention, huggingface2024mastering, chen2023survey,keles2023computational,liu2022pyraformer}. These limitations present major obstacles to deploying LMMs in resource-constrained or latency-sensitive environments \cite{chen2024fastv,lin2025vtw}.

Recent studies have shown that visual token representations in LMMs exhibit substantial redundancy \cite{liu2024mustdrop,shang2024prumerge,huang2024ivtp,tong2025flowcut,li2025todre}. Leveraging this insight, visual token pruning methods have been proposed to reduce computational cost by selectively removing unnecessary tokens. By eliminating redundant visual tokens, these approaches effectively alleviate the quadratic burden of long input sequences. Notably, prior work demonstrates that pruning the majority of visual tokens can yield significant efficiency gains with minimal performance degradation \cite{zhang2024fastervlm,chen2024fastv,lin2025vtw,huang2024ivtp,sunvelar}.

While token pruning is beneficial, its application to LMMs remains challenging. Existing token pruning methods can be broadly categorized into three types. 1) A common approach is to use attention scores to identify redundant tokens \cite{chen2024image,lin2025boosting,shang2024llava,tong2025flowcut}. However, such methods are susceptible to positional bias and often retain spatially adjacent tokens with high similarity, leading to redundancy and performance degradation. 2) Other methods rely on model-specific calibration or fine-tuning \cite{lin2025boosting,li2025tokenpacker,cai2024matryoshka}, which incurs high computational cost and limits scalability in practical deployment. 3) Another line of work addresses token pruning by maximizing the minimum distance between tokens \cite{alvar2025divprune}. While this strategy encourages token separation, it is sensitive to outliers and may fail to preserve cross-modal alignment, ultimately affecting downstream performance.

% While the advantages of token pruning are well recognized, applying it to LMMs remains non-trivial. A line of methods leverages attention scores to identify redundant tokens; however, these approaches often suffer from positional bias and tend to preserve highly similar tokens, leading to redundancy and suboptimal performance. Other methods require costly fine-tuning or calibration, which hinders scalability and practical deployment. Difference-based pruning typically relies on held-out calibration sets and assumes model-specific distribution alignment, making it less flexible when adapting to new LMMs backbones. Similarity-based strategies, which merge redundant tokens into the most similar retained ones, have been empirically shown to be inferior to direct pruning. Additionally, methods that aim to maximize the minimum pairwise distance are sensitive to outliers and frequently neglect cross-modal alignment, further compromising their effectiveness. Beyond these limitations, existing work has largely overlooked the inherent differences in redundancy between visual and textual tokens.
\begin{figure}[t]
    \centering
    \includegraphics[width=1\linewidth]{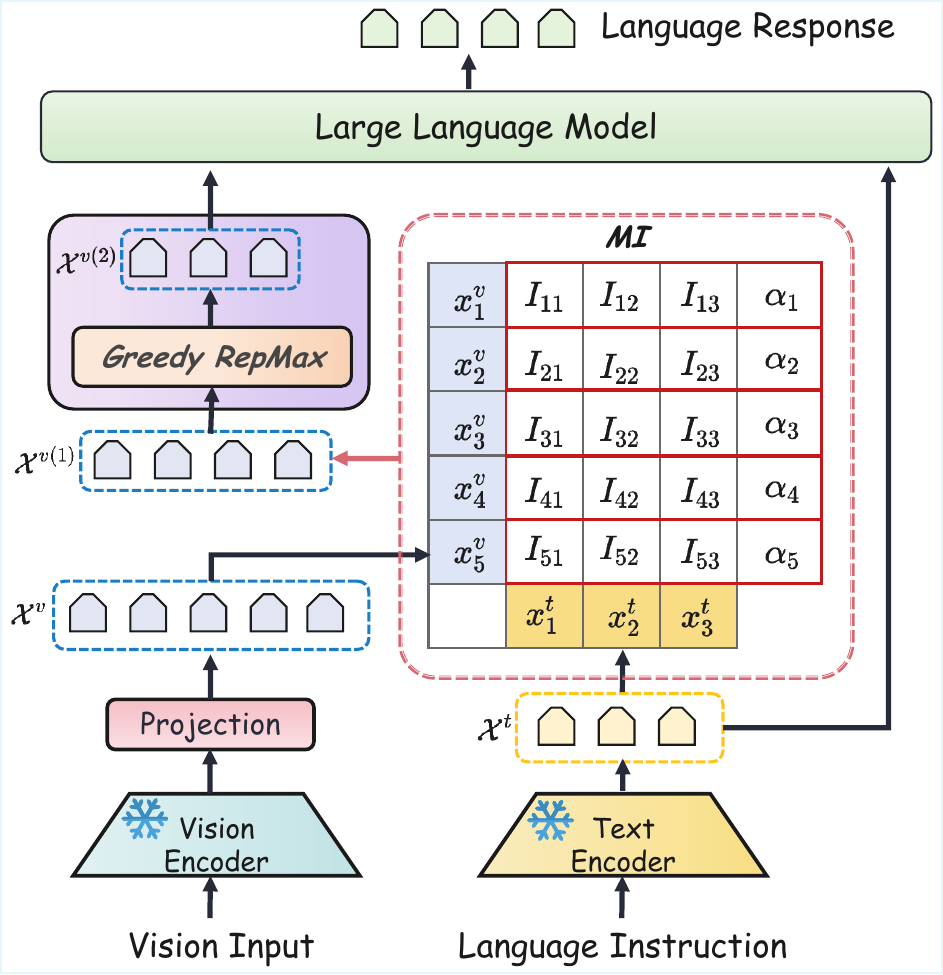}
    \caption{Overview of the visual token pruning method. We compute the average mutual information (MI) between each visual token and textual tokens to obtain its semantic alignment score $\alpha_i$, and preserve the highest score tokens to form $\mathcal{X}^{v(1)}$. This subset is further refined by Greedy RepMax, which prunes redundant tokens to yield $\mathcal{X}^{v(2)}$. Greedy RepMax is a greedy approximation to the {NP Hard} problem of maximizing the expected pairwise distance among visual tokens.}
    \label{fig:fig1_res}
\end{figure}

In view of these challenges and opportunities, we propose a visual token pruning strategy that explicitly preserves both cross-modal alignment and intra-modal informational diversity. Our analysis begins by examining the inherent redundancy differences between visual and textual tokens through three key perspectives: attention mechanisms, semantic distribution, and information repetition. These analyses reveal that visual tokens exhibit significantly higher redundancy compared to their textual counterparts. Consequently, we choose to prune only visual tokens while retaining all textual tokens. To preserve cross-modal alignment during pruning, we estimate the mutual information between visual and textual token embeddings. Visual tokens with higher expected mutual information with textual tokens are more likely to exhibit semantic alignment with the textual modality and are therefore retained, while those with lower mutual information are pruned as misaligned. Additionally, we retain visual tokens by maximizing their expected pairwise distances in the embedding space, thereby promoting intra-modal information diversity and encouraging rich, non-overlapping visual semantic representations. To address this problem efficiently, we employ a greedy algorithm that iteratively selects tokens exhibiting maximal dissimilarity. Extensive experiments demonstrate that our method maintains strong performance while reducing tokens by 88.9\% on models such as LLaVA-1.5-7B and LLaVA-NEXT-7B, resulting in a 56.7\% improvement in inference speed. In summary, our contributions can be summarized as follows:
\begin{itemize}
  \item We analyze the redundancy differences between visual and textual tokens and propose pruning exclusively on visual tokens. Based on this, we design a visual token pruning method that explicitly preserves cross-modal alignment and intra-modal information diversity, improving inference efficiency while maintaining semantic integrity.
  \item We introduce a mutual information based pruning strategy that retains semantically aligned visual tokens with textual tokens while removing misaligned ones, effectively preserving the alignment between visual and textual modalities.
  \item We propose an information diversity driven pruning strategy that maximizes the expected pairwise distances among visual tokens in the embedding space, effectively reducing redundancy and enhancing intra modal information richness, which is solved efficiently with a greedy algorithm.
\end{itemize}

% In this paper, we propose a visual token pruning method that explicitly preserves both cross-modal alignment and intra-modal diversity. We begin by analyzing the inherent redundancy differences between visual and textual tokens from multiple perspectives, and consequently decide to prune only visual tokens while keeping all textual tokens intact. To retain cross-modal alignment, we compute the mutual information between visual and textual token representations. Visual tokens with higher expected mutual information with text tokens are more likely to be semantically aligned with the language modality, while those with lower mutual information are considered misaligned and pruned accordingly. To further enhance the diversity of the retained visual tokens, we maximize the mathematical expectation of the pairwise distances among them in the vector space, thereby ensuring rich and non-redundant visual representations. Comprehensive experiments across a wide range of models and datasets validate the effectiveness and generalizability of our approach. 

\section{Related Work}
% \subsection{Large Multimodal Models (LMMs)}
% Large Multimodal Models (LMMs) have enhanced the reasoning capabilities of pretrained large language models (LLMs) by enabling joint comprehension of visual and textual modalities, including images and videos. Typically, LMMs employ a vision encoder to extract dense visual features, which are then projected into the embedding space of the language model via modules such as Q-Former or multilayer perceptrons. Conventional approaches often resize high-resolution images to a fixed size, introducing geometric distortions and degrading fine-grained spatial details. To mitigate these limitations, dynamic tiling techniques partition images into smaller regions, encoding each independently with a shared encoder. Although this strategy better preserves local information, it significantly increases the number of visual tokens, leading to greater computational demands. This challenge is further amplified in video-based LMMs, where processing multiple frames compounds the token count and inference complexity. These factors underscore the need for efficient inference methods to facilitate the deployment of LMMs in resource-constrained settings.
\subsection{Large Multimodal Models (LMMs)}
{Large Multimodal Models (LMMs)} extend pretrained LLMs by integrating multiple modalities \cite{bai2025qwen25vl,team2025kimivl}, such as images and texts. Typically, a vision encoder extracts dense visual features, which are projected into the language model’s embedding space via modules like Q-Former \cite{chebotar2023q} or MLPs \cite{taud2017multilayer}. Conventional methods resize high-resolution images to a fixed scale \cite{koonce2021resnet}, causing geometric distortions and loss of fine-grained spatial details. Dynamic tiling mitigates this by splitting images into smaller regions, each independently encoded by a shared encoder, better preserving local information \cite{yuan2021tokens,yin2022vit}. However, this increases the number of visual tokens and computational cost, a challenge further amplified in video LMMs due to multi-frame processing \cite{liu2024mustdrop}. These challenges highlight the need for efficient inference methods to enable LMMs deployment under resource constraints \cite{tong2025flowcut}.

\subsection{Visual Token Pruning}
{Visual Token Pruning} aims to reduce computational overhead and improve inference efficiency by removing redundant or less informative visual tokens. A common strategy is to leverage attention scores to identify tokens for removal. For example, PruMerge \cite{shang2024llava} clusters and merges tokens in the vision encoder based on attention sparsity, while FastV \cite{chen2024fastv} utilizes attention weights from the second layer of the LLM to guide pruning. SparseVLM \cite{zhang2024sparsevlm} employs cross-modal attention for text-conditioned token selection, and VisionZip \cite{yang2025visionzip} compresses visual inputs via CLS token attention in the final vision encoder layer. FlowCut \cite{tong2025flowcut} identifies redundancy from an information flow perspective by analyzing attention propagation. LVPruning \cite{sun2025lvpruning} uses cross-attention to assess vision token importance via interaction with language tokens, guiding token pruning. Other methods adopt calibration-based strategies, where pruning ratios and depths are determined by evaluating model behavior on a held-out set. FitPrune \cite{ye2025fitprune} compares attention distributions before and after pruning to inform token selection, while VTW \cite{lin2025vtw} shows that tokens can be safely dropped after specific layers, guided by calibration. In addition, approaches like DivPrune \cite{alvar2025divprune} addresses token pruning by maximizing the minimum distance between tokens. Moreover, a concurrent work \cite{zhang2025beyond} reformulates the token pruning problem using determinantal point processes to achieve effective pruning.

% \begin{figure*}[t]
%     \centering
%     \includegraphics[width=1\linewidth]{1.pdf}
%     \caption{Overview of the visual token pruning method. Visual tokens that exhibit low semantic alignment with textual tokens are pruned by measuring their cross-modal mutual information. The remaining tokens are further reduced by maximizing the intra-modal representation within the visual modality, which is efficiently solved using a greedy algorithm.}
%     \label{fig:fig1}
% \end{figure*}

\section{Token Redundancy Across Modalities}
\textbf{Previous studies \cite{shang2024prumerge,alvar2025divprune,li2025todre} primarily decide to prune visual tokens based on the discrepancy in the number of visual and textual tokens. However, these works do not thoroughly analyze the differences in redundancy across modalities.} In this work, we provide a more detailed pruning guidance by examining the redundancy of visual and textual tokens from \textbf{three perspectives: attention mechanisms, semantic distribution and information repetition}. Our findings reveal that visual tokens exhibit substantially higher redundancy than textual tokens. Motivated by this observation, we retain all textual tokens during the pruning process and apply pruning exclusively to visual tokens. The detailed analysis can be found in Appendix \ref{appendix_redundacncy}.

\section{Methodology}
\subsection{Problem Formulation}
In this section, we propose a visual token pruning strategy that explicitly preserves cross-modal alignment while promoting intra-modal information diversity. Given an input image, we extract a sequence of visual tokens \(\mathcal{X}^v = \{x_1^v, \ldots, {x}_ {N}^v\}\), and embed the accompanying text into a sequence of textual tokens \(\mathcal{X}^t = \{{x}_1^t, \ldots, {x}_M^t\}\), where \({x}_i^v, {x}_j^t \in \mathbb{R}^d\). 

Our pruning pipeline can be roughly divided into two stages. In the first stage, we select a subset \(\mathcal{X}^{v(1)} \subset \mathcal{X}^v\) with size \(\left|\mathcal{X}^{v(1)}\right| = {N}_1\), where \({N}_1 < N\), by retaining tokens with the highest semantic alignment to the text modality, thereby ensuring the preservation of tokens with strong cross-modal correlations. In the second stage, we further prune \(\mathcal{X}^{v(1)}\) to obtain a more compact subset \(\mathcal{X}^{v(2)} \subset \mathcal{X}^{v(1)}\) with size \(\left|\mathcal{X}^{v(2)}\right| = {N}_2\), where \({N}_2 < {N}_1\), by maximizing the expected pairwise distance in the embedding space. This promotes non-redundant and information-rich representations while reducing computational overhead. We provide the experimental analysis of the stage order and the used metrics in Section~\ref{5_5_Ablation} and Appendix~\ref{appendix_ablation}. Figure \ref{fig:fig1_res} illustrates the overview of our method.

\subsection{Cross-Modal Alignment Aware Token Filtering}
During training, multimodal models primarily optimize cross-entropy loss for text generation, which often leads to an implicit weakening of cross-modal alignment \cite{covert2024locality}. Pruning as a post-processing step for the model is crucial for cross-modal alignment. To preserve cross-modal alignment during visual token pruning, we introduce a mutual information based criterion that selects a subset of visual tokens \(\mathcal{X}^{v(1)}\) which maximizes the expected mutual information with the textual tokens. Our selection objective can be formulated as follows:
\begin{equation}
\mathcal{X}^{v(1)} 
= \mathop{\arg\max}_{\substack{\mathcal{X}^{v'} \subset \mathcal{X}^v, \\ |\mathcal{X}^{v'}| = N_1}} 
\mathbb{E}_{x^t \in \mathcal{X}^t} \left[ I(\mathcal{X}^{v'}; x^t) \right]
\end{equation}
Here, \(\mathcal{X}^{v'}\) serves as a temporary variable. \(I(\mathcal{X}^{v'}; x^t)\) denotes the average mutual information between the textual token \(x^t\) and all visual tokens in the subset \(\mathcal{X}^{v'}\). Furthermore, we define the alignment score \(\alpha_i\) of each visual token \(x_i^v\) based on its mutual information with the textual token set \(\mathcal{X}^t\) as follows:
\begin{equation}
\alpha_i = I(x_i^v; \mathcal{X}^t) = \frac{1}{M} \sum_{j=1}^M I(x_i^v; {x}_j^t)
\end{equation}
This score reflect the average semantic alignment strength of visual tokens within the shared embedding space. Intuitively, tokens with higher alignment scores are retained, while those with lower semantic alignment are pruned.

Although mutual information provides a theoretically grounded measure of cross-modal alignment, its precise estimation in high dimensional spaces often requires costly density modeling. We approximate mutual information using the \( L_2 \) norm, which reduces computational overhead without compromising performance. It is worth noting that our approximation is grounded in mathematical theory and does not rely on any specific model architecture. \textbf{ Additionally, we provide theoretical justification and experimental analysis to explain why the \( L_2 \) norm can be used for approximation (see Appendix \ref{proof} and \ref{appendix_ablation}).} Therefore, we can compute \( \alpha_i \) using the following formula:
\begin{equation}
\alpha_i = - \frac{1}{M} \sum_{j=1}^M \|{x}_i^v - {x}_j^t\|_2
\end{equation}

% When considering token embeddings \( x_i^v \) and \( x_j^t \), the point mutual information  can be defined as follows:
% \begin{equation}
% I_p(x_i^v; x_j^t) = \log \frac{p(x_i^v \mid x_j^t)}{p(x_i^v)}.
% \end{equation}
% where \(I_p\) represents point mutal information. We assume that the conditional distribution \( p(x_i^v \mid x_j^t) \) follows an isotropic Gaussian distribution, a common assumption inspired by the InfoNCE ~\cite{rusak2024infonce} and the variational inference framework used in VAEs \cite{davidson2018hyperspherical}. Under this assumption, we have:
% \begin{equation}
% p(x_i^v \mid x_j^t) = \mathcal{N}(x_j^t, \sigma^2 I),
% \end{equation}
% \begin{equation}
% \log p(x^v \mid x^t) = C_1 - \frac{1}{2\sigma^2} \| x^v - x^t \|_2^2, \ C_1 = - \frac{d}{2} \log (2 \pi \sigma^2)
% \end{equation}
% where \( C_1 \) is a constant independent of the inputs. The marginal distribution \( p(x^v) \) is generally unknown, but in the approximation it can be treated as a constant (denoted as \( C_2 \)).the pointwise mutual information becomes:
% \begin{equation}
% I_p(x_i^v; x_j^t) \approx -\frac{1}{2\sigma^2} \|x_i^v - x_j^t\|_2^2 + (C_1 - C_2),
% \end{equation}
% Therefore, maximizing mutual information reduces to minimizing the squared Euclidean distance between the embeddings \( x_i^v \) and \( x_j^t \), which provides a theoretical justification for using Euclidean distance as a surrogate objective under the Gaussian assumption.

We then construct the set \(\mathcal{X}^{v(1)}\) by selecting the \({N_1}\) visual tokens with the highest alignment scores, which can be represented as follows:
% \begin{equation}
% \mathcal{X}^{v(1)} = \left\{{x}_i^v \in \mathcal{X}^v \mid \alpha_i \in top_{N_1}(\alpha_1, \alpha_2, \ldots, \alpha_N)\right\}
% \end{equation}
\begin{equation}
\scalebox{0.87}{$
\mathcal{X}^{v(1)} = \left\{{x}_i^v \in \mathcal{X}^v \mid 
\alpha_i \in top_{N_1}(\alpha_1, \alpha_2, \ldots, \alpha_N)\right\}
$}
\end{equation}
This selection effectively filters out semantically misaligned visual tokens while retaining those with the strongest cross-modal correlations, thereby enhancing alignment quality and reducing computational overhead. Even for semantically sparse queries (e.g., image captioning on the COCO dataset), our method consistently achieves strong performance.

\paragraph{Superiority over Other Works}
Some works prune visual tokens based on their attention scores with textual tokens. However, these approaches are susceptible to positional bias, often retaining spatially adjacent tokens that are highly redundant. Although both our method and these approaches leverage interactions between text and vision, our method emphasizes alignment in the semantic space rather than relying on positional correlations. Moreover, we do not rely on an additional CLIP model to remap the embeddings, as this would alter the vector space and introduce extra computational overhead. We provide a detailed comparison under different metrics in Appendix~\ref{appendix_ablation}, where the results show that the $L_2$ norm achieves better performance compared to other measures.

\subsection{Greedy Intra-Modal Representational Maximization (RepMax)}
After obtaining the visual token set \(\mathcal{X}^{v(1)}\), we select a refined subset \(\mathcal{X}^{v(2)}\) by explicitly maximizing intra-modal information diversity. The core idea is to retain a set of visual tokens that are semantically diverse and non-redundant. To achieve this, we maximize the expected pairwise distance among the selected tokens. Specifically, we define the distance between two visual tokens \({x}_i^v\) and \({x}_j^v\) using cosine dissimilarity. Based on this, we formulate the objective for subset selection as follows:
\begin{equation}
D({x}_i^v, {x}_j^v) = 1 - \frac{({x}_i^v)^\top {x}_j^v}{\|{x}_i^v\|_2 \cdot \|{x}_j^v\|_2}
\end{equation}
\begin{equation}
\begin{aligned}
\mathcal{X}^{v(2)}=\mathop{\arg\max}_{\substack{\widetilde{\mathcal{X}} \subset \mathcal{X}^{v(1)}, \\ |\widetilde{\mathcal{X}}| = N_2}}  \mathbb{E}_{\substack{{x}_i^v, {x}_j^v \in \widetilde{\mathcal{X}}, i \neq j}} 
\left[ D({x}_i^v, {x}_j^v) \right]
\end{aligned}
\end{equation}

% \begin{equation}
% \mathcal{X}^{v(2)} = \arg\max_{\widetilde{\mathcal{X}} \subset \mathcal{X}^{v(1)},\, |\widetilde{\mathcal{X}}| = N_2} \; \mathbb{E}_{{x}_i^v, {x}_j^v \in \widetilde{\mathcal{X}},\, i \neq j} \left[ D({x}_i^v, {x}_j^v) \right]
% \end{equation}
% \begin{equation}
% \mathcal{X}^{v(2)} = \arg\max_{\widetilde{\mathcal{X}} \subset \mathcal{X}^{v(1)},\, |\widetilde{\mathcal{X}}| = N_2} \; \mathbb{E}_{{x}_i^v, {x}_j^v \in \widetilde{\mathcal{X}}} \left[ D({x}_i^v, {x}_j^v) \right]
% \end{equation}
This objective encourages the model to retain visual tokens that capture complementary aspects of the input, thereby enhancing the representational richness of the pruned token set. To address the problem, we introduce a binary selection vector \(\boldsymbol{\gamma} \in \{0,1\}^{N_1}\). The objective can be expressed as follows:
% \begin{equation}
% \max_{\boldsymbol{\gamma} \in \{0,1\}^{N_1}} \sum_{i=1}^{N_1} \sum_{\substack{j=1 , j \neq i}}^{N_1} \gamma_i \gamma_j \cdot D({x}_i^v, {x}_j^v) \\ , \quad \sum_{i=1}^{N_1} \gamma_i = N_2
% \end{equation}
% \begin{equation}
% \max\limits_{\boldsymbol{\gamma} \in \{0,1\}^{N_1}}
% \sum\limits_{\substack{i=1, j=1 \\ i \neq j}}^{N_1}
% \gamma_i \gamma_j \cdot D(x_i^v, x_j^v), \; \sum_{i=1}^{N_1} \gamma_i = N_2
% \end{equation}
\begin{equation}
\scalebox{0.92}{$
\max\limits_{\boldsymbol{\gamma} \in \{0,1\}^{N_1}}
\sum\limits_{\substack{i=1, j=1 \\ i \neq j}}^{N_1}
\gamma_i \gamma_j \cdot D(x_i^v, x_j^v),
\;\;
\sum\limits_{i=1}^{N_1} \gamma_i = N_2
$}
\end{equation}

The above problem is a combinatorial optimization task, which is known to be NP Hard. To address this, we adopt an efficient greedy algorithm for approximate optimization. The core idea is to iteratively select the next visual token that is farthest from the current selected subset in the embedding space, thereby progressively enhancing overall information diversity. Our experimental results further validate the effectiveness of the greedy algorithm.

At the initial stage of the greedy algorithm, we quantify pairwise relationships among visual tokens by constructing a cosine similarity matrix \(\mathcal{C} \in \mathbb{R}^{N_1 \times N_1}\). Each element \(\mathcal{C}_{ij}\) of the matrix \(\mathcal{C}\) represents the similarity between tokens \(x_i^v\) and \(x_j^v\). Based on this matrix, we compute the average similarity of each token with all others. The token with the lowest average similarity is chosen as the initial seed to initialize the selected set for the first iteration (\(\mathcal{S}^{(1)}\)), and the remaining tokens constitute the initial remaining set (\(\mathcal{R}^{(1)}\)). This can be formulated as follows:
\begin{equation}
    \mathcal{C}_{ij} = \frac{({x}_i^v)^\top {x}_j^v}{\|{x}_i^v\|_2 \cdot \|{x}_j^v\|_2}
\end{equation}
\begin{equation}
    c^{(1)}_i = \frac{1}{N_1} \sum_{j=1}^{N_1} \mathcal{C}_{ij}, \quad \forall i \in [N_1]
\end{equation}
\begin{equation}
    s^{(1)} = \arg\min_{i \in [N_1]} c^{(1)}_i
\end{equation}
\begin{equation}
    \mathcal{S}^{(1)} = \{s^{(1)} \},\; \mathcal{R}^{(1)} = [N_1] \setminus \{s^{(1)}\}
\end{equation}
where \( c^{(1)}_i \) denotes the average similarity between token \( {x}^v_i \) and the other \( N_1 \) visual tokens. The index \( s^{(1)} \) corresponds to the initial token selected as the least similar on average to all others, and \( [N_1] \) denotes the set of indexs from 1 to \( N_1 \). To enable efficient incremental computation, we maintain a similarity accumulation vector \({\sigma}  \in \mathbb{R}^{N_1} \), where each element \(\sigma_i\) denotes the average similarity between token \({x}_i^v\) and all tokens in the selected set. The initial similarity accumulation vector \( {\sigma}^{(1)} \) is given by the similarity between the token indexed by \( s^{(1)} \) and all other tokens, which can be represented as follows: 
\begin{equation}
    \sigma^{(1)} = \mathcal{C}_{s^{(1)}:} 
\end{equation}
where \( \mathcal{C}_{s^{(1)}:} \) denotes the row vector in the matrix \( \mathcal{C} \) corresponding to the index \( s^{(1)} \).

At each iteration \( t = 2, \ldots, N_2 \), we compute the average similarity between each remaining token and all tokens in the current selected set. The remaining token with the lowest average similarity is selected in this iteration, and its index is used to update both the selected and remaining sets. Meanwhile, the total similarity accumulation vector is incrementally updated to incorporate the newly selected token. This process is formalized as follows:
\begin{equation}
    {c}_i^{(t)} = \frac{1}{t - 1} \cdot \sigma^{(t - 1)}_i, \quad \forall i \in \mathcal{R}^{(t - 1)}
\end{equation}
\begin{equation}
    s^{(t)} = \arg\min_{i \in \mathcal{R}^{(t - 1)}} {c}_i^{(t)}
\end{equation}
% \begin{equation}
%     \mathcal{S}^{(t)} = \mathcal{S}^{(t - 1)} \cup \{s^{(t)}\},  \mathcal{R}^{(t)} = \mathcal{R}^{(t - 1)} \setminus \{s^{(t)}\}
% \end{equation}
\begin{equation}
\scalebox{0.91}{$
\mathcal{S}^{(t)} = \mathcal{S}^{(t - 1)} \cup \{s^{(t)}\},\;
\mathcal{R}^{(t)} = \mathcal{R}^{(t - 1)} \setminus \{s^{(t)}\}
$}
\end{equation}
\begin{equation}
    \sigma^{(t)} = \sigma^{(t - 1)} + \mathcal{C}_{s^{(t)}:}
\end{equation}
Here, \( \mathcal{S}^{(t)} \) and \( \mathcal{R}^{(t)} \) denote the selected and remaining sets after the \( t \)-th iteration, respectively. \( {c}_i^{(t)} \) represents the average similarity between the token \( x_i^v \) in the remaining set and all tokens in the selected set at iteration \( t \). The index \( s^{(t)} \) corresponds to the token selected during the \( t \)-th iteration, and \( \mathcal{C}_{s^{(t)}:} \) denotes the row vector in the matrix \( \mathcal{C} \) corresponding to the index \( s^{(t)} \). The similarity accumulation vector \( {\sigma}^{(t)} \) is updated using the similarity vector \(\mathcal{C}_{s^{(t)}:}\) of the newly selected token.

This process is repeated \( N_2 \) times until \( N_2 \) tokens have been selected. The resulting set \(\mathcal{X}^{v(2)} \subset \mathcal{X}^{v(1)}\) forms a maximally dispersed subset of visual tokens with minimal internal redundancy.

\paragraph{Superiority over Other Works}
We compare our method with DivPrune, which performs pruning by maximizing the minimum distance. In our experiments, we observe that outliers in the token embeddings can significantly affect DivPrune, preventing it from achieving optimal performance. In contrast, our method optimizes the expected pairwise distance, which effectively mitigates the influence of outliers, and employs a designed greedy algorithm to efficiently tackle the associated NP hard problem. The results of our comparative experiments demonstrate a substantial performance improvement over DivPrune. Furthermore, our method maintains strong performance when compared to an extended variant (DivPrune$^{*}$) implemented by our method (see Section~\ref{appendix_DivPrune*}).

\begin{table*}[t]
\centering
\resizebox{\textwidth}{!}{
\begin{tabular}{c|cccccccccc}
\toprule
\textbf{Method} & \textbf{ChartQA} & \textbf{COCO} & \textbf{MMB} & \textbf{MME} & \textbf{MMU} & \textbf{NoCaps} & \textbf{OCRB} & \textbf{VQA\textsubscript{OK}} & \textbf{POPE} & \textbf{VQA\textsubscript{TEXT}} \\ \midrule
\multicolumn{11}{c}{\textit{Upper Bound, 576 Tokens $(100\%)$}} \\
Dense & 18.20 & 1.10 & 64.09 & 1508.24 & 36.33 & 1.06 & 31.30 & 53.44 & 93.86 & 46.11 \\ \midrule
\multicolumn{11}{c}{\textit{Retain 192 Tokens $(\downarrow 66.7\%)$}} \\
ToMe & 15.45 & 0.08 & 58.64 & 1262.74 & 31.49 & 0.08 & 29.19 & 46.88 & 77.64 & 39.60 \\
FastV & 15.24 & 0.08 & 63.06 & 1302.33 & 31.26 & 0.09 & 29.53 & 47.24 & 69.49 & 38.43 \\
SpVLM & 17.17 & 1.05 & 61.62 & 1390.39 & 35.55 & 1.00 & 30.16 & 50.47 & 89.66 & 42.00 \\
PDrop & 17.06 & 1.06 & 60.41 & 1426.75 & 34.42 & 0.99 & 30.35 & 50.83 & 88.26 & 41.78 \\
VisionZip & 17.48 & 1.06 & 62.92 & 1440.02 & 35.88 & 1.01 & 30.41 & 51.14 & 92.20 & 43.01 \\
DivPrune & 17.40 & 1.06 & 62.29 & 1436.44 & 35.78 & 1.01 & 30.40 & 51.55 & 91.37 & 43.24 \\
\textbf{Ours} & \textbf{17.80} & \textbf{1.08} & \textbf{63.75} & \textbf{1467.30} & \textbf{36.22} & \textbf{1.03} & \textbf{31.00} & \textbf{52.29} & \textbf{93.68} & \textbf{45.10} \\ \midrule
\multicolumn{11}{c}{\textit{Retain 128 Tokens $(\downarrow 77.8\%)$}} \\
ToMe & 14.13 & 0.06 & 57.27 & 1066.04 & 30.69 & 0.06 & 24.93 & 43.24 & 68.92 & 37.90 \\
FastV & 13.86 & 0.07 & \textbf{62.89} & 1182.73 & 31.71 & 0.08 & 26.24 & 44.56 & 65.41 & 35.88 \\
SpVLM & 16.56 & 1.04 & 60.08 & 1346.25 & 33.85 & 1.01 & 28.06 & 48.35 & 88.35 & 40.51 \\
PDrop & 16.35 & 1.03 & 59.97 & 1320.85 & 34.76 & 1.01 & 28.81 & 49.85 & 90.32 & 41.30 \\
VisionZip & 17.00 & 1.05 & 61.94 & 1393.33 & 36.10 & 1.02 & 28.88 & 49.66 & 93.75 & 42.71 \\
DivPrune & 16.96 & 1.04 & 61.77 & 1396.25 & 36.22 & \textbf{1.04} & 29.00 & 50.02 & 91.31 & 41.66 \\
\textbf{Ours} & \textbf{17.80} & \textbf{1.06} & 62.80 & \textbf{1397.22} & \textbf{36.44} & 1.02 & \textbf{29.50} & \textbf{51.12} & \textbf{94.29} & \textbf{44.28} \\ \midrule
\multicolumn{11}{c}{\textit{Retain 64 Tokens $(\downarrow 88.9\%)$}} \\
ToMe & 12.49 & 0.05 & 49.97 & 909.99 & 27.46 & 0.06 & 20.07 & 39.47 & 62.85 & 34.59 \\
FastV & 12.03 & 0.05 & 58.94 & 1004.35 & 29.40 & 0.06 & 22.04 & 41.65 & 57.46 & 32.81 \\
SpVLM & 14.92 & 0.99 & 58.03 & 1203.46 & 26.55 & 0.95 & 25.81 & 45.14 & 89.91 & 37.51 \\
PDrop & 15.14 & 0.99 & 45.42 & 1248.24 & 34.98 & 0.94 & 27.00 & 44.09 & 66.92 & 33.81 \\
VisionZip & 15.74 & 1.00 & 59.91 & 1348.03 & 35.80 & 0.95 & 28.50 & 48.85 & 92.85 & 41.31 \\
DivPrune & 15.84 & 0.99 & 59.28 & 1348.99 & 35.89 & 0.94 & 27.60 & 48.36 & 92.18 & 39.22 \\
\textbf{Ours} & \textbf{16.36} & \textbf{1.02} & \textbf{61.34} & \textbf{1359.12} & \textbf{36.11} & \textbf{0.98} & \textbf{29.00} & \textbf{49.37} & \textbf{95.01} & \textbf{41.45} \\
\bottomrule
\end{tabular}
}
\caption{Comparison of visual token pruning methods on LLaVA-1.5-7B across multiple benchmarks under varying token retention ratios. Best results are highlighted in bold.}
\label{tab:t1}
\end{table*}

\section{Experiments}
\subsection{Experimental Setup}
\textbf{Models and Baselines. }
We evaluate the performance of our method across several representative LMMs, including LLaVA-1.5-7B \cite{liu2023llava1.5}, LLaVA-1.5-13B \cite{liu2023llava1.5}, LLaVA-NEXT-7B \cite{liu2024llavanext}, LLaVA-NEXT-Video-7B \cite{zhang2024llavanextvideo} and Qwen2-VL-7B \cite{wang2024qwen2}. These models span a range of parameter scales and encompass both image and video understanding tasks. We evaluate our method alongside several widely adopted baselines, including ToMe \cite{bolya2022tome}, FastV \cite{chen2024fastv}, SparseVLM (SpVLM) \cite{zhang2024sparsevlm}, PDrop \cite{xing2024pdrop}, VisionZip \cite{yang2025visionzip} and DivPrune \cite{alvar2025divprune}. We additionally include the VTW \cite{lin2025boosting} method in our efficiency evaluation. All baselines are assessed under consistent experimental settings.

\textbf{Datasets and Metrics. }We evaluate our method on diverse multimodal benchmarks covering various tasks. For ChartQA \cite{masry2022chartqa}, we report the relaxed score. Image captioning quality and diversity are measured by CIDEr on COCO \cite{sharma2018conceptual} and NoCaps \cite{agrawal2019nocaps}. MMBench\textsubscript{EN} (MMB) \cite{liu2024mmbench} uses a GPT-based score. Perceptual understanding is assessed via the perception score on MME \cite{Fu2023mme}. Accuracy evaluates general reasoning on MMU \cite{zheng2025multi} and text recognition on OCRBench \cite{liu2024ocrbench}. Exact match accuracy measures question answering on VQA\textsubscript{OK} \cite{marino2019ok} and VQA\textsubscript{TEXT} \cite{singh2019textvqa}. POPE \cite{li2023pope} is evaluated with precision for positional understanding. This comprehensive protocol ensures thorough assessment of generalization across multimodal tasks.

\textbf{Implementation Details. }{Our experiments are conducted using the PyTorch framework \cite{paszke2019pytorch} and the Hugging Face Transformers library \cite{wolf2020transformers}. We utilize an NVIDIA H100 GPU with 80GB of memory.} We set the parameter \( N_1 = \left\lfloor 0.8N \right\rceil
 \)  by default. In both the ablation study and case study, we fix the final number of retained tokens to 64. The temperature, prompt and preprocessing in our experiments all follow the default settings of {LMMs-Eval}.

\begin{table}[ht]
\centering
\resizebox{0.48\textwidth}{!}{
\begin{tabular}{c|cccccc}
\toprule
\textbf{Method} & \textbf{ChartQA} & \textbf{COCO} & \textbf{MMB} & \textbf{NoCaps} & \textbf{OCRB} & \textbf{POPE} \\ \midrule
 & \multicolumn{6}{c}{\textit{Upper Bound, 576 Tokens $(100\%)$}} \\
Dense & 18.20 & 1.16 & 68.73 & 1.09 & 33.60 & 94.44 \\ \midrule
 & \multicolumn{6}{c}{\textit{Retain 144 Tokens $(\downarrow 75.0\%)$}} \\
VisionZip & \textbf{17.44} & \textbf{1.10} & 65.76 & 1.03 & 31.19 & 95.54 \\
DivPrune & 17.28 & 1.09 & 66.41 & 1.02 & 31.10 & 94.63 \\
\textbf{Ours} & \textbf{17.44} & \textbf{1.10} & \textbf{67.35} & \textbf{1.04} & \textbf{33.00} & \textbf{96.69} \\ \midrule
 & \multicolumn{6}{c}{\textit{Retain 128 Tokens $(\downarrow 77.8\%)$}} \\
VisionZip & 17.09 & \textbf{1.09} & 66.02 & 1.02 & 31.01 & 95.04 \\
DivPrune & 17.08 & \textbf{1.09} & 66.07 & 1.01 & 30.90 & 94.76 \\
\textbf{Ours} & \textbf{17.88} & \textbf{1.09} & \textbf{66.84} & \textbf{1.04} & \textbf{33.10} & \textbf{96.55} \\ \midrule
 & \multicolumn{6}{c}{\textit{Retain 64 Tokens $(\downarrow 88.9\%)$}} \\
VisionZip & 16.29 & 1.04 & 64.44 & 0.97 & 29.77 & 95.03 \\
DivPrune & 16.32 & 1.03 & 64.52 & 0.97 & 30.10 & 95.84 \\
\textbf{Ours} & \textbf{16.56} & \textbf{1.05} & \textbf{64.95} & \textbf{0.99} & \textbf{30.60} & \textbf{97.63} \\
\bottomrule
\end{tabular}%
}
\caption{Comparison between our visual token pruning method and DivPrune on LLaVA-1.5-13B across multiple benchmarks under varying token retention settings.}
\label{tab:llava-1.5-13b}
\end{table}

\subsection{Main Results}
We conduct a comprehensive evaluation of our visual token pruning approach against several baselines. As shown in Table~\ref{tab:t1}, our method consistently outperforms prior works across a range of token retention ratios on LLaVA-1.5-7B. At a moderate retention of 192 tokens, our approach achieves an MME perception score of 1467.30, with only a 2.7\% drop relative to the dense model, compared to 13.7\% and 16.2\% drops for FastV and ToMe, respectively. Compared with the strongest baseline, DivPrune, our method improves MMB accuracy from 62.28 to 63.75, a relative gain of 2.4\%. To further assess the scalability of our approach on larger models, we extend our evaluation to LLaVA-1.5-13B and conduct a comparative analysis against VisionZip and DivPrune under varying visual token retention settings. As presented in Table~\ref{tab:llava-1.5-13b}, our method consistently surpasses others, demonstrating its effectiveness at higher model scale. These results highlight the generalization ability of our pruning strategy in large scale settings.

In addition, we conduct experiments on LLaVA-NEXT-7B using a larger pool of visual tokens across different retention settings. As shown in Table \ref{tab:llava-16-7b}, our method consistently achieves the strongest performance across a range of compression levels. Furthermore, our experiments on LLaVA-NEXT-Video-7B and Qwen2-VL-7B also demonstrate the effectiveness of the proposed method (see Appendix \ref{appendix_qwen}).

\begin{table}[htbp]
\centering
\resizebox{0.48\textwidth}{!}{
\begin{tabular}{c|cccccc}
\toprule
\textbf{Method} & \textbf{ChartQA} & \textbf{COCO} & \textbf{MME} & \textbf{NoCaps} & \textbf{OCRB} & \textbf{POPE} \\ \midrule
 & \multicolumn{6}{c}{\textit{Upper Bound, 2880 Tokens $(100\%)$}} \\
Dense & 54.88 & 1.00 & 1519.30 & 0.88 & 52.50 & 95.71 \\ \midrule
 & \multicolumn{6}{c}{\textit{Retain 720 Tokens $(\downarrow 75.0\%)$}} \\
ToMe & 36.57 & 0.08 & 1294.12 & 0.09 & 42.25 & 81.58 \\
FastV & 36.09 & 0.09 & 1334.69 & 0.10 & 42.74 & 73.01 \\
SpVLM & 40.66 & 0.95 & 1424.94 & 0.82 & 43.65 & 94.20 \\
PDrop & 40.39 & 0.95 & 1462.20 & 0.83 & 43.93 & 92.73 \\
VisionZip & 40.91 & 0.95 & 1480.45 & 0.84 & 44.07 & 96.27 \\
DivPrune & 41.20 & 0.95 & 1472.13 & 0.83 & 44.00 & 96.00 \\
\textbf{Ours} & \textbf{44.92} & \textbf{1.00} & \textbf{1503.24} & \textbf{0.86} & \textbf{46.40} & \textbf{96.44} \\ \midrule
 & \multicolumn{6}{c}{\textit{Retain 640 Tokens $(\downarrow 77.8\%)$}} \\
ToMe & 32.43 & 0.08 & 1133.13 & 0.08 & 37.05 & 72.73 \\
FastV & 31.82 & 0.08 & 1257.15 & 0.09 & 39.00 & 69.02 \\
SpVLM & 37.99 & 0.94 & 1430.96 & 0.83 & 41.71 & 93.22 \\
PDrop & 37.53 & 0.94 & 1403.96 & 0.82 & 42.82 & 95.31 \\
VisionZip & 39.01 & 0.96 & 1475.23 & 0.83 & 42.87 & 96.67 \\
DivPrune & 38.92 & 0.96 & 1484.12 & 0.83 & 43.10 & 96.35 \\
\textbf{Ours} & \textbf{43.72} & \textbf{0.99} & \textbf{1492.60} & \textbf{0.99} & \textbf{45.10} & \textbf{96.76} \\ \midrule
 & \multicolumn{6}{c}{\textit{Retain 320 Tokens $(\downarrow 88.9\%)$}} \\
ToMe & 25.27 & 0.08 & 976.13 & 0.08 & 25.23 & 66.17 \\
FastV & 24.34 & 0.08 & 1077.35 & 0.08 & 27.71 & 60.50 \\
SpVLM & 30.18 & 0.92 & 1290.93 & 0.80 & 32.45 & 94.66 \\
PDrop & 30.62 & 0.92 & 1338.97 & 0.80 & 33.95 & 70.46 \\
VisionZip & 32.14 & 0.94 & 1445.85 & 0.81 & 35.35 & 95.71 \\
DivPrune & 32.04 & 0.93 & 1447.05 & 0.80 & 34.70 & \textbf{97.05} \\
\textbf{Ours} & \textbf{35.64} & \textbf{0.96} & \textbf{1452.37} & \textbf{0.82} & \textbf{37.90} & {96.47} \\
\bottomrule
\end{tabular}
}
\caption{Performance comparison of our visual token pruning method with other baselines on LLaVA-NEXT-7B across multiple benchmarks under varying token retention settings (retaining 720, 640, and 320 tokens).}
\label{tab:llava-16-7b}
\end{table}

\subsection{Efficiency Analysis (Accounting for Pruning Overhead)}
In this section, we evaluate the inference efficiency of our method on LLaVA-NEXT-7B using the MME dataset. We repeat the process three times and report the average result. For a fair comparison, the number of decoding steps for each method is fixed to the minimum decoding steps required among all methods. As shown in Table~\ref{tab:latency}, we report the average inference time (time to generate the complete output) per sample with a batch size of 1 and 288 visual tokens retained. Compared to the original dense model (15.53 GB, 235.2 ms), all pruning methods significantly reduce memory usage and cut inference latency by more than half. Our method achieves an inference time of 100.13 ms, representing a 57\% reduction compared to the dense baseline. When compared with other pruning baselines, our approach achieves lower latency than VTW and FastV. Although our latency is slightly higher than DivPrune, this small gap is negligible given the superior task performance consistently achieved under the same token budget. It is worth noting that pruning is performed during inference; therefore, the reported inference latency includes the overhead introduced by the pruning process. These results demonstrate that our pruning strategy provides substantial improvements in memory and latency efficiency.

\begin{table}[]
\centering
\begin{tabular}{c|cc}
\toprule
\textbf{Method} & \textbf{Memory (G)} & \textbf{Latency (ms)} \\ \midrule
Dense & 15.53 & 235.2 \\
VTW & 13.63 & 103.71 \\
FastV & 13.63 & 108.29 \\
DivPrune & 13.63 & {99.53} \\
Ours & {13.63} & 100.13 \\ \bottomrule
\end{tabular}
\caption{Comparison of peak GPU memory usage and average per-sample latency for our pruning method and baselines on LLaVA-NEXT-7B, evaluated on the MME dataset with a batch size of 1 and 288 retained tokens.}
\label{tab:latency}
\end{table}

% \begin{table*}[htbp]
% \centering
% \setlength{\tabcolsep}{2.5pt}
% \begin{tabular}{c|cccccccccc}
% \toprule
% \textbf{\( N_1 \)} & \textbf{ChartQA} & \textbf{COCO} & \textbf{MMB} & \textbf{MME} & \textbf{MMU} & \textbf{NoCaps} & \textbf{OCRB} & \textbf{VQA\textsubscript{OK}} & \textbf{POPE} & \textbf{VQA\textsubscript{TEXT}} \\
% \midrule
% \( \left\lfloor 0.9N \right\rceil \)  & 16.48 & 1.02 & 60.99 & 1332.41 & 36.11 & 0.97 & 29.00 & 49.25 & 95.20 & 41.60 \\
% \( \left\lfloor 0.8N \right\rceil \)  & 16.36 & 1.02 & 61.34 & 1339.12 & 36.11 & 0.98 & 29.00 & 49.37 & 95.01 & 41.45 \\
% \( \left\lfloor 0.75N \right\rceil \) & 16.28 & 1.02 & 61.08 & 1334.68 & 36.11 & 0.97 & 28.70 & 48.93 & 95.43 & 40.78 \\
% \( \left\lfloor 0.7N \right\rceil \)  & 16.00 & 1.02 & 61.43 & 1347.34 & 35.78 & 0.97 & 27.80 & 48.43 & 95.31 & 39.03 \\
% \bottomrule
% \end{tabular}%
% \caption{Ablation study on the effect of different filtering ratios in Cross-Modal Alignment Aware Token Filtering on LLaVA-v1.5-7B.}
% \label{tab:stage1_ration}
% \end{table*}

\subsection{Comparison with our DivPrune variants (DivPrune$^{*}$)}
\label{appendix_DivPrune*}
We further extend the DivPrune approach to enable a more comprehensive comparison. Concretely, we integrate our cross-modal pruning strategy with DivPrune under its default configuration, forming a new variant denoted as \text{DivPrune*}. We evaluate this variant on LLaVA-1.5-7B while retaining 64 visual tokens. As shown in Table~\ref{tab:comparison1}, our method still surpasses \text{DivPrune*}. Combined with the earlier results, we also observe that \text{DivPrune*} achieves noticeable gains over the original DivPrune, highlighting the crucial role of cross-modal alignment in boosting performance. Moreover, these findings demonstrate that our greedy intra-modal expectation representation maximization strategy is more effective than the minimum distance maximization employed in DivPrune.
\begin{table}[htbp]
\centering
\resizebox{0.48\textwidth}{!}{
\begin{tabular}{c|cccccc}
\toprule
\textbf{Method} & \textbf{ChartQA} & \textbf{COCO} & \textbf{MMB} & \textbf{MME} & \textbf{MMU} & \textbf{NoCaps} \\
\midrule
DivPrune$^{*}$ & 15.98 & 1.01 & 60.28 & 1355.28 & 36.02 & 0.96 \\
Ours       & 16.36 & 1.02 & 61.34 & 1359.12 & 36.11 & 0.98 \\
\bottomrule
\end{tabular}%
}
\caption{Comparison of DivPrune variants (DivPrune$^{*}$) with our method on LLaVA-1.5-7B across multiple benchmarks.}
\label{tab:comparison1}
\end{table}

\subsection{Ablation Study}
\label{5_5_Ablation}
To further validate the effectiveness of our approach, we conduct ablation studies with three different configurations. $ablation_1$ swaps the order of the two pruning stages by applying intra-modal representation maximization before cross-modal alignment. $ablation_2$ removes intra-modal pruning and only retains cross-modal alignment based pruning. $ablation_3$ eliminates cross-modal alignment pruning and applies only intra-modal representation maximization. As shown in Table~\ref{many_ablation}, the performance consistently drops once any component is removed, indicating the necessity of each part of our framework. Moreover, using only cross-modal pruning or altering the pruning order leads to more severe degradation. We conjecture that this is because text-related visual information is often concentrated in local regions, where tokens align with the text but may carry highly similar semantics. Additional experiments in Section~\ref{appendix_hyper} further demonstrate that with properly chosen pruning parameters for each stage, our method achieves strong performance.

\begin{table}[htbp]
\centering
\resizebox{0.48\textwidth}{!}{
\begin{tabular}{c|cccccc}
\toprule
\textbf{Method} & \textbf{ChartQA} & \textbf{COCO} & \textbf{MMB} & \textbf{MME} & \textbf{MMU} & \textbf{NoCaps} \\
\midrule
$ablation_1$ & 13.12 & 0.83 & 50.00 & 1114.77 & 34.56 & 0.76 \\
$ablation_2$ & 13.08 & 0.83 & 49.57 & 1108.64 & 34.33 & 0.75 \\
$ablation_3$ & 16.16 & 1.00 & 59.97 & 1311.46 & 36.02 & 0.97 \\
Ours         & 16.36 & 1.02 & 61.34 & 1359.12 & 36.11 & 0.98 \\
\bottomrule
\end{tabular}%
}
\caption{Ablation study of different pruning strategies on LLaVA-1.5-7B across multiple benchmarks.}
\label{many_ablation}
\end{table}

\subsection{Additional Results}
We conduct a \textbf{Hyperparameter Analysis}, such as varying \(N_1\), and observe consistently strong performance across different settings, demonstrating the robustness of our method (Appendix \ref{appendix_hyper}). Moreover, we perform a \textbf{Case Study} to show that the pruned model retains its ability to generate high quality content (Appendix \ref{appendix_case}).

\section{Conclusion}
In this paper, we analyze the redundancy discrepancy between visual and textual tokens in LMMs and propose a pruning strategy that operates exclusively on visual tokens. Our method explicitly preserves cross-modal alignment and intra-modal informational diversity. Specifically, we leverage mutual information to eliminate visual tokens that are semantically misaligned with textual inputs, ensuring the consistency across modalities. To further enhance the representational quality of the retained tokens, we maximize their expected pairwise distances in the embedding space via an efficient greedy algorithm. Extensive experiments across diverse models and benchmarks demonstrate the effectiveness of our approach, alongside significant improvements in memory consumption and inference latency. We plan to extend our approach to more complex or non-visual modalities in future work, further demonstrating its broad applicability.

\section*{Limitations}
In this work, we conduct extensive experiments to evaluate the effectiveness of our multimodal model token pruning method. The results demonstrate that our approach achieves competitive performance compared to the baselines. However, due to computational constraints, we have not yet been able to evaluate it on larger scale models, such as those with 70 billion parameters. Exploring the scalability of our method to such large models constitutes an important direction for future work.


\bibliography{custom}

\appendix
\newpage
\newpage

\section{Token Redundancy Across Modalities}
\label{appendix_redundacncy}
{Previous studies \cite{shang2024prumerge,alvar2025divprune,li2025todre} primarily decide to prune visual tokens based on the discrepancy in the number of visual and textual tokens. However, these works do not thoroughly analyze the differences in redundancy across modalities.} In this work, we provide a more detailed pruning guidance by examining the redundancy of visual and textual tokens from {three perspectives: attention mechanisms, semantic distribution and information repetition}. Our findings reveal that visual tokens exhibit substantially higher redundancy than textual tokens. Motivated by this observation, we retain all textual tokens during the pruning process and apply pruning exclusively to visual tokens.

The attention distribution in multimodal models is highly skewed, with textual tokens receiving significantly higher weights than visual tokens \cite{li2025vista, lee2025tamp}. This suggests that the model primarily relies on text for semantic understanding. Visual inputs often consist of hundreds or thousands of tokens, far exceeding the number of textual tokens. Given the quadratic complexity of attention with respect to token count, this mismatch between quantity and contribution is pronounced. Despite the large number of visual tokens, their impact on the final representation is limited, resulting in poor trade offs between computational cost and semantic gain and reflecting the high redundancy of visual tokens.

Each token in text typically carries explicit semantic meaning, such as a noun, verb, or conjunction, reflecting natural language as a highly optimized discrete encoding system with dense and relatively uniform information distribution \cite{hirschberg2015advances}. In contrast, semantic information in images is highly concentrated in a small number of salient regions, such as foreground objects, while the majority of visual tokens correspond to low-semantic areas like sky or walls, which mostly contain low-frequency textures or superficial variations and contribute little to high-level semantic understanding. Consequently, the effective semantic density of visual tokens is significantly lower than that of textual tokens.

Compared to textual tokens, visual tokens exhibit higher information redundancy due to differences in their generation processes and underlying information structures. Textual tokens, derived from natural language, follow Zipf’s law \cite{newman2005power, adamic2002zipf} and represent distinct semantic units with low repetition. In contrast, visual tokens are typically generated by uniformly partitioning images or extracting low-level features without semantic organization. For example, a sky region is divided into multiple similar patches that cluster in embedding space and jointly represent a single high-level concept, causing redundancy. Visual tokenization resembles physical partitioning rather than frequency-driven semantic abstraction, forcing models to process many similar tokens and resulting in computational and representational inefficiency.

\section{Proof of \(L_2\) Norm Approximation to Mutual Information}
\label{proof}
When considering two continuous random variables \( x_i^v  \) and \( x_j^t \), corresponding to two distributions, the mutual information between them is defined as follows:
% \begin{equation}
% I(x_i^v; x_j^t) = \int p(x_i^v, x_j^t) \log \frac{p(x_i^v \mid x_j^t)}{p(x_i^v)} \, dx_i^v dx_j^t
% \end{equation}
\begin{equation}
\scalebox{0.99}{$
I(x_i^v; x_j^t) 
= \int p(x_i^v, x_j^t) 
\log \frac{p(x_i^v \mid x_j^t)}{p(x_i^v)} 
\, dx_i^v dx_j^t
$}
\end{equation}

To render this quantity tractable, we make the following assumptions:
\begin{itemize}
    \item The conditional distribution \( p(x_i^v \mid x_j^t) \) is modeled as an \textit{isotropic Gaussian}. We conduct extensive experiments and find that, during the fitting process, the conditional distribution exhibits per-dimension means and standard deviations concentrated around (-0.0035) and (0.7842), respectively, which substantiates the validity of our hypothesis. Then, we can derive the following expression:
    \begin{equation}
    p(x_i^v \mid x_j^t) = \mathcal{N}(x_j^t, \sigma^2 I)
    \end{equation}
    Here, \( I \) denotes the identity matrix.
    \item The marginal distribution \( p(x_i^v) \) is generally unknown and intractable; thus, we approximate it as a constant \( C_2 \), following common practice in variational inference and contrastive learning.
\end{itemize}
Based on the above assumptions, we obtain the following expression:
\begin{equation}
\log p(x_i^v \mid x_j^t) = C_1 - \frac{1}{2\sigma^2} \| x_i^v - x_j^t \|_2^2
\end{equation}
% \begin{equation}
% I(x_i^v; x_j^t) 
% &\approx -\frac{1}{2\sigma^2} \, \mathbb{E}_{ p(x_i^v, x_j^t)} \left[ \| x_i^v - x_j^t \|_2^2 \right] + C
% \end{equation}
\begin{equation}
\scalebox{0.93}{$
I(x_i^v; x_j^t) 
\approx -\frac{1}{2\sigma^2} \, 
\mathbb{E}_{p(x_i^v, x_j^t)} \!\left[ \| x_i^v - x_j^t \|_2^2 \right] + C
$}
\end{equation}
Here, \( C_1 = -\frac{d}{2} \log(2\pi\sigma^2) \) is a constant independent of the inputs, and we denote \( C = C_1 - C_2 \). Therefore, the \( L_2 \) norm provides a theoretically grounded approximation of mutual information.

% In contrast, although cosine similarity is widely used in contrastive learning for its simplicity, it is inherently unsuitable as a proxy for mutual information. It captures only angular relationships and ignores feature magnitudes, allowing low-norm, potentially uninformative tokens to yield high similarity scores. It captures only angular relationships while ignoring feature magnitudes, which allows tokens with low norms and potentially little semantic information to produce high scores.
% Moreover, cosine similarity lacks a probabilistic foundation and typically requires very strong assumptions to establish any connection to mutual information.

% \paragraph{Intra-Modal Maximization.}
% To further assess the effectiveness of our intra-modal representation maximization strategy, we conduct an ablation study that takes the combination of maximizing minimum intra-modal distance and cross-modal alignment as a strong baseline. Specifically, we perform experiments on LLaVA-1.5-7B while retaining 64 visual tokens. As presented in Table \ref{tab:comparison1}, our method consistently outperforms the baseline, clearly demonstrating its effectiveness.

\section{Additional Comparative Experiments}
\label{appendix_qwen}
We evaluate our method on the video based multimodal model LLaVA-NEXT-Video-7B. Using the same pruning strategy as in our image understanding experiments, we conduct evaluations on the COCO dataset and report standard image captioning metrics, including BLEU-1/2/3/4, ROUGE-L and CIDEr. We compare our method against DivPrune, a really strong baseline for visual token pruning. As shown in Table~\ref{tab:llava-video-7b}, our approach consistently outperforms DivPrune across all evaluation metrics under various token retention settings. For instance, with 960 tokens retained, our method improves the BLEU-4 score from 28.46 to 30.41, representing a relative gain of 6.9\%. These results demonstrate the robustness of our method in video scenarios.
\begin{table}[htbp]
\centering
\resizebox{0.48\textwidth}{!}{
\begin{tabular}{c|cccccc}
\toprule
\textbf{Method} & \textbf{B-1} & \textbf{B-2} & \textbf{B-3} & \textbf{B-4} & \textbf{R-L} & \textbf{CIDEr} \\ \midrule
 & \multicolumn{6}{c}{\textit{Upper Bound, 2880 Tokens $(100\%)$}} \\
Dense & 71.07 & 54.78 & 40.50 & 29.24 & 53.58 & 1.02 \\ \midrule
 & \multicolumn{6}{c}{\textit{Retain 960 Tokens $(\downarrow 66.7\%)$}} \\
DivPrune & 72.51 & 54.97 & 40.00 & 28.46 & 53.79 & 0.99 \\
\textbf{Ours} & \textbf{73.27} & \textbf{56.63} & \textbf{42.03} & \textbf{30.41} & \textbf{54.67} & \textbf{1.04} \\ \midrule
 & \multicolumn{6}{c}{\textit{Retain 720 Tokens $(\downarrow 75.0\%)$}} \\
DivPrune & 72.35 & 54.81 & 39.88 & 28.37 & 53.61 & 0.99 \\
\textbf{Ours} & \textbf{72.87} & \textbf{56.34} & \textbf{41.77} & \textbf{30.23} & \textbf{54.47} & \textbf{1.04} \\ \midrule
 & \multicolumn{6}{c}{\textit{Retain 540 Tokens $(\downarrow 81.3\%)$}} \\
DivPrune & 71.72 & 54.16 & 39.21 & 27.75 & 53.28 & 0.98 \\
\textbf{Ours} & \textbf{72.34} & \textbf{55.55} & \textbf{40.92} & \textbf{29.47} & \textbf{53.75} & \textbf{1.03} \\ \midrule
 & \multicolumn{6}{c}{\textit{Retain 480 Tokens $(\downarrow 83.3\%)$}} \\
DivPrune & 71.28 & 53.72 & 38.88 & 27.47 & 53.13 & 0.97 \\
\textbf{Ours} & \textbf{72.44} & \textbf{55.49} & \textbf{40.74} & \textbf{29.21} & \textbf{53.80} & \textbf{1.01} \\
\bottomrule
\end{tabular}
}
\caption{Performance comparison of our visual token pruning method and DivPrune on LLaVA-NEXT-Video-7B, evaluated on COCO using standard captioning metrics. B denotes BLEU, and R denotes ROUGE.}
\label{tab:llava-video-7b}
\end{table}

To more comprehensively assess the effectiveness of our method, we further perform pruning experiments on Qwen2-VL-7B, pruning 88.9\% of tokens. We test our approach against VisionZip and {DivPrune} across multiple datasets. As illustrated in Table \ref{Qwen}, our method achieves superior performance, underscoring its robustness and effectiveness even under extreme pruning scenarios.
\begin{table}[htbp]
\centering

\begin{tabular}{c|ccc}
\toprule
\textbf{Method} & \textbf{MMB} & \textbf{MME} & \textbf{VQA\textsubscript{TEXT}} \\
\midrule
Dense    & 78.26  & 1894 & 65 \\
VisionZip & 74.26 & 1702.97 & 57.18 \\
DivPrune & 73.96  & 1699.03 & 58.29 \\
Ours     & {74.90}  & {1706.74} & {58.43} \\
\bottomrule
\end{tabular}
\caption{Performance comparison of 88.9\% token pruning on Qwen2-VL-7B.}
\label{Qwen}
\end{table}

\section{Metric Analysis}
\label{appendix_ablation}
In addition to the theoretical analysis of approximating mutual information (MI) using the \( L_2 \) norm, we conduct empirical validation on LLaVA-1.5-7B with 64 visual tokens retained. Specifically, we implement cross-modal alignment using three metrics: mutual information, its \( L_2 \) norm approximation, and cosine similarity. We use scikit-learn \cite{pedregosa2011scikit} to compute mutual information, where the library employs a nonparametric \(k\) nearest neighbors approximation method, with \(k\) set to 3 by default. As shown in Table \ref{tab:metric_comparison2}, the method using the \( L_2 \) norm approximation achieves performance comparable to that of exact mutual information, and consistently outperforms the one using cosine similarity. We guess that this phenomenon occurs because semantic alignment between textual and visual tokens requires not only directional consistency but also a substantial overlap in their feature distributions. We further compare performance under different intra-modal redundancy metrics, namely cosine similarity and the \(L_2\) norm. As shown in Table~\ref{unimodal}, cosine similarity consistently yields better results. This suggests that cosine similarity is generally more effective for measuring redundancy within unimodal data, which is consistent with prior empirical findings in the literature.
\begin{table}[htbp]
\centering
\resizebox{0.48\textwidth}{!}{
\begin{tabular}{c|cccccc}
\toprule
\textbf{Metric} & \textbf{ChartQA} & \textbf{COCO} & \textbf{MMB} & \textbf{MME} & \textbf{MMU} & \textbf{NoCaps} \\
\midrule
MI   & 16.38 & 1.02 & 61.38 & 1362.88 & 36.12 & 0.98 \\
\(L_2\) & 16.36 & 1.02 & 61.34 & 1359.12 & 36.11 & 0.98 \\
cos  & 16.26 & 1.01 & 61.15 & 1357.05 & 35.83 & 0.97 \\
\bottomrule
\end{tabular}%
}
\caption{Performance comparison among various cross-modal alignment metrics across multiple benchmarks.}
\label{tab:metric_comparison2}
\end{table}

\begin{table}[htbp]
\centering
\resizebox{0.48\textwidth}{!}{
\begin{tabular}{c|cccccc}
\toprule
\textbf{Metric} & \textbf{ChartQA} & \textbf{COCO} & \textbf{MMB} & \textbf{MME} & \textbf{MMU} & \textbf{NoCaps} \\
\midrule
\(L_2\) & 16.32 & 1.0 & 61.26 & 1356.91 & 36.02 & 0.98 \\
cos & 16.36 & 1.02 & 61.34 & 1359.12 & 36.11 & 0.98  \\
\bottomrule
\end{tabular}%
}
\caption{Performance comparison among various intra-modal redundancy metrics across multiple benchmarks.}
\label{unimodal}
\end{table}

\section{Hyperparameter Analysis} 
\label{appendix_hyper}
In this study, we conduct an analysis of the influence of the token retention parameter \( N_1 \) in the first stage cross-modal token pruning process on the overall performance of our method. As presented in Table~\ref{tab:stage1_ration}, we evaluate several representative configurations on LLaVA-1.5-7B, including \( N_1 = \left\lfloor 0.9N \right\rceil \), \( N_1 = \left\lfloor 0.8N \right\rceil \), \( N_1 = \left\lfloor 0.75N \right\rceil \), and \( N_1 = \left\lfloor 0.7N \right\rceil \), where \( N \) represents the total number of visual tokens in the dense model. In all cases, the final number of retained tokens is fixed at 64. Our method consistently achieves strong and stable performance across these different settings, indicating that it is largely insensitive to the choice of \( N_1 \). These strong results highlight the robustness of our method to different hyperparameter settings.

\begin{table}[htbp]
\centering
\resizebox{0.48\textwidth}{!}{
\begin{tabular}{c|cccccc}
\toprule
\textbf{\( N_1 \)} & \textbf{ChartQA} & \textbf{COCO} & \textbf{MMB} & \textbf{MME} & \textbf{MMU} & \textbf{NoCaps} \\
\midrule
\( \left\lfloor 0.9N \right\rceil \)  & 16.48 & 1.02 & 60.99 & 1332.41 & 36.11 & 0.97 \\
\( \left\lfloor 0.8N \right\rceil \)  & 16.36 & 1.02 & 61.34 & 1359.12 & 36.11 & 0.98 \\
\( \left\lfloor 0.75N \right\rceil \) & 16.28 & 1.02 & 61.08 & 1334.68 & 36.11 & 0.97 \\
\( \left\lfloor 0.7N \right\rceil \)  & 16.00 & 1.02 & 61.43 & 1347.34 & 35.78 & 0.97 \\
\bottomrule
\end{tabular}%
}
\caption{Impact of different values of the parameter \( N_1 \) on model performance using LLaVA-1.5-7B.}
\label{tab:stage1_ration}
\end{table}

\section{Case Study}
\label{appendix_case}
\begin{figure*}[tbh!]
    \centering
    \begin{subfigure}{\linewidth}
        \centering
        \includegraphics[width=0.91\textwidth]{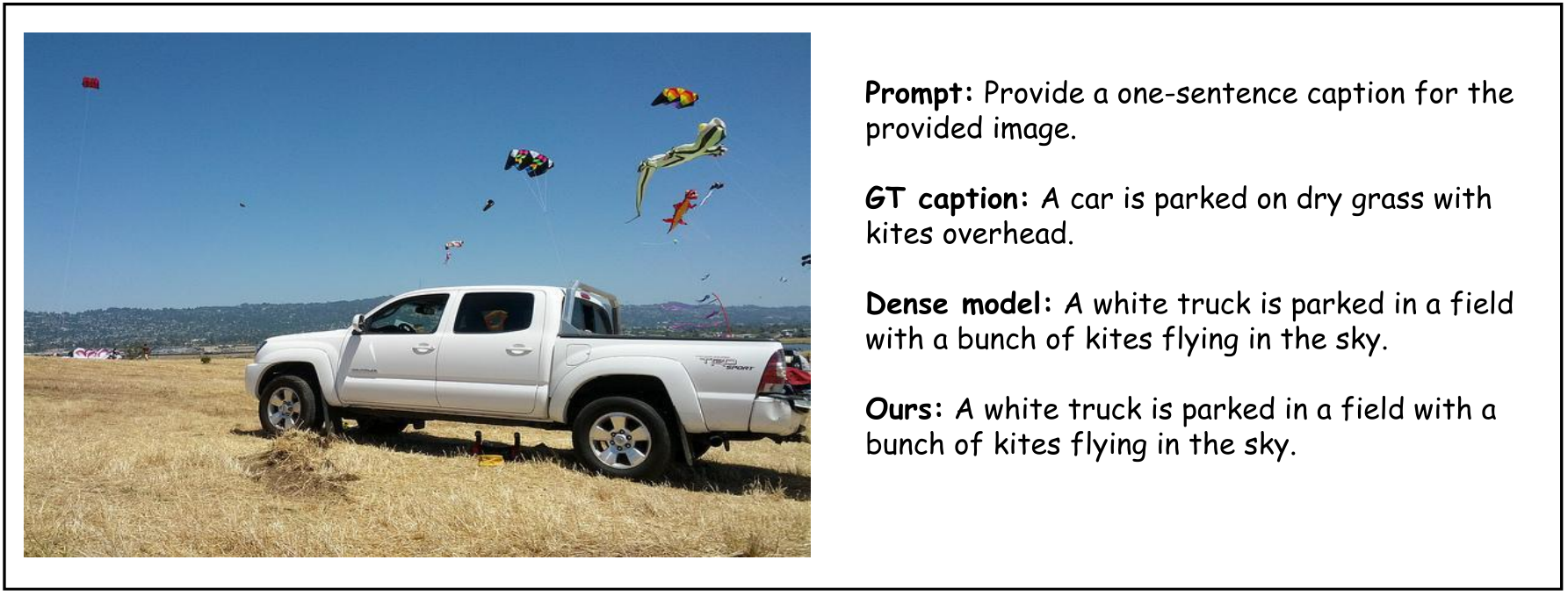}
        \caption{}
        %\label{fig:figure1}
    \end{subfigure}
     
    \begin{subfigure}{\linewidth}
        \centering
        \includegraphics[width=0.91\textwidth]{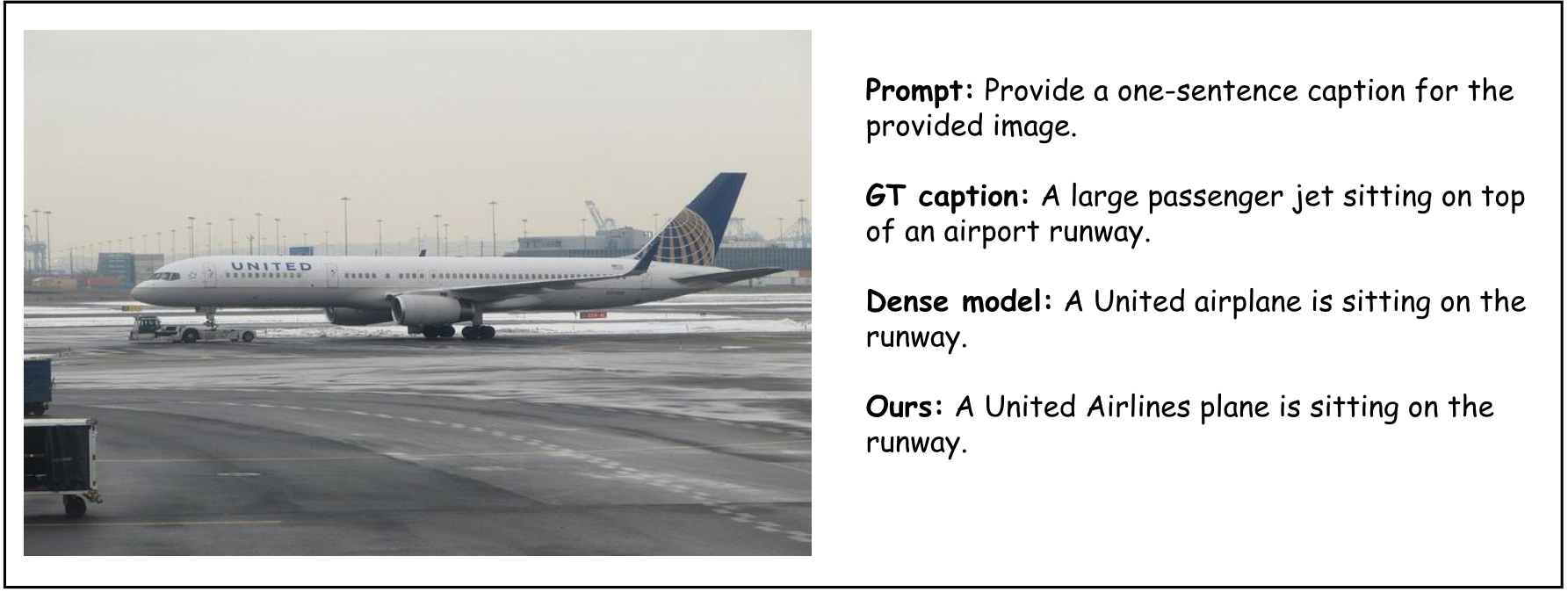}        
        \caption{}
        %\label{fig:figure2}
    \end{subfigure}

    \begin{subfigure}{\linewidth}
        \centering
        \includegraphics[width=0.91\textwidth]{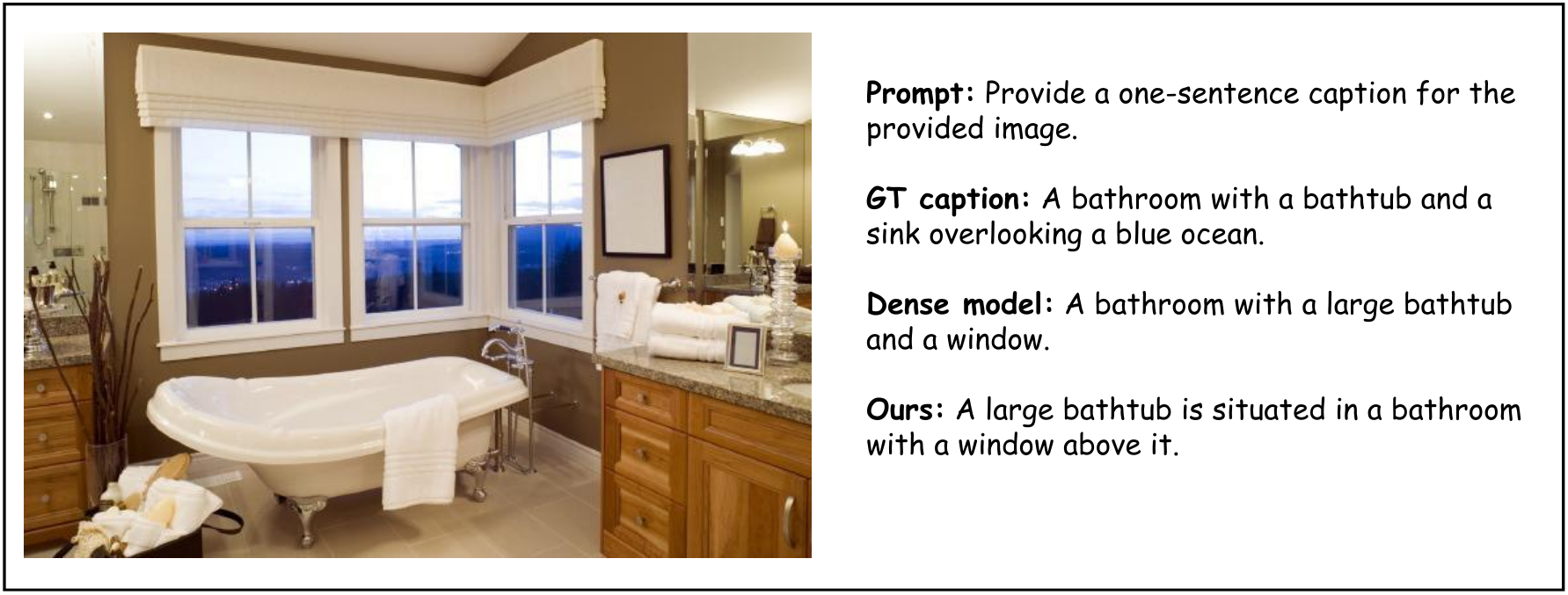}        
        \caption{}
        %\label{fig:figure3}
    \end{subfigure}
    \caption{Case study comparing captions generated by dense and pruned models (LLaVA-1.5-7B) on the COCO dataset, demonstrating output consistency despite substantial token reduction.}
    \label{fig:case_study}
\end{figure*}

To qualitatively assess the effectiveness of our pruning strategy, we conduct a case study using LLaVA-1.5-7B on samples from the COCO dataset. As illustrated in Figure~\ref{fig:case_study}, we compare the captions generated by the original dense model and our pruned model, which retains significantly fewer visual tokens. The results show that the pruned model produces captions that remain semantically coherent and fluent, closely aligning with those generated by the dense model. This demonstrates that our method effectively removes redundant visual tokens while preserving caption generation capabilities. Overall, the case study highlights that substantial token reduction can be achieved without compromising output quality, underscoring the practical utility of our approach.

\section*{Reproducibility statement}
We strive to ensure the reproducibility of our results. Full details are provided in the main paper and the appendix. Our implementation is built on PyTorch and standard open-source libraries. We provide key code implementations to facilitate reproducibility and further research.

\end{document}